\documentclass[journal]{IEEEtran}
\usepackage{cite}
\usepackage{amsmath,amssymb,amsfonts}
\usepackage{algorithmicx}
\usepackage{graphicx}
\usepackage{textcomp}
\usepackage{xcolor}
\usepackage{multirow}
\usepackage{booktabs}
\usepackage{subcaption}
\usepackage{hyperref}
\usepackage{todonotes}
\usepackage{color,soul}
\usepackage[ruled,vlined]{algorithm2e}

\SetKwInput{KwInput}{Input}

\hypersetup{
colorlinks = true,
linkcolor = red,
anchorcolor = blue,
citecolor = green,
filecolor = magenta,
urlcolor = cyan
}

\ifCLASSINFOpdf

\else

\fi

\begin{document}

\title{LDNet: End-to-End Lane Marking Detection Approach Using a Dynamic Vision Sensor}

\author{Farzeen Munir,~\IEEEmembership{Student Member,~IEEE,}
Shoaib~Azam,~\IEEEmembership{Student Member,~IEEE,}
Moongu~Jeon,~\IEEEmembership{Senior Member,~IEEE,} Byung-Geun Lee,~\IEEEmembership{Member,~IEEE,} and Witold Pedrycz,~\IEEEmembership{Life Fellow, IEEE} 
\thanks{Farzeen~Munir, Shoaib~Azam, Byung-Geun Lee  and Moongu~Jeon are
with the School of Electrical Engineering and Computer Science,
Gwangju Institute of Science and Technology,
Gwangju, South Korea.
e-mail:((farzeen.munir, shoaibazam, mgjeon, bglee)@gist.ac.kr) 
\newline
W. Pedrycz is with the Department of Electrical and Computer Engineering,
University of Alberta, Edmonton, AB T6R 2V4, Canada, with the Department
of Electrical and Computer Engineering, Faculty of Engineering, King Abdulaziz University, Jeddah 21589, Saudi Arabia, and also with the Systems Research Institute, Polish Academy of Sciences, Warsaw 01-447, Poland. 
\newline
e-mail:(wpedrycz@ualberta.ca)
\newline
This work was partly supported by the ICT R$\&$D program of MSIP/IITP (2014-3-00077, Development of global multitarget tracking and event prediction techniques based on real-time large-scale video analysis), National Research Foundation of Korea (NRF) grant funded by the Korea Government (MSIT) (No. 2019R1A2C2087489), Ministry of Culture, Sports and Tourism (MCST), and Korea Creative Content Agency (KOCCA) in the Culture Technology (CT) Research \& Development (R2020070004) Program 2021.}}

\maketitle

\begin{abstract}
Modern vehicles are equipped with various driver-assistance systems, including automatic lane keeping, which prevents unintended lane departures. Traditional lane detection methods incorporate handcrafted or deep learning-based features followed by postprocessing techniques for lane extraction using frame-based RGB cameras. The utilization of frame-based RGB cameras for lane detection tasks is prone to illumination variations, sun glare, and motion blur, which limits the performance of lane detection methods. Incorporating an event camera for lane detection tasks in the perception stack of autonomous driving is one of the most promising solutions for mitigating challenges encountered by frame-based RGB cameras. The main contribution of this work is the design of the lane marking detection model, which employs the dynamic vision sensor. This paper explores the novel application of lane marking detection using an event camera by designing a convolutional encoder followed by the attention-guided decoder. The spatial resolution of the encoded features is retained by a dense atrous spatial pyramid pooling (ASPP) block. The additive attention mechanism in the decoder improves performance for high dimensional input encoded features that promote lane localization and relieve postprocessing computation. The efficacy of the proposed work is evaluated using the DVS dataset for lane extraction (DET). The experimental results show a significant improvement of $5.54\%$ and $5.03\%$ in $F1$ scores in multiclass and binary-class lane marking detection tasks. Additionally, the intersection over union ($IoU$) scores of the proposed method surpass those of the best-performing state-of-the-art method by $6.50\%$ and $9.37\%$ in multiclass and binary-class tasks, respectively.

\end{abstract}


\IEEEpeerreviewmaketitle

\section{Introduction}
%
%
%
%
\IEEEPARstart{A}{dvancements} in the development of sensor technology have made a tremendous impact on autonomous driving in terms of environmental perception  \cite{Bengler2014}.  In the context of autonomous vehicles, the architecture mainly comprises a sensor layer, perception layer, planning layer, and control layer \cite{swn}. The sensor layer includes the integration of exteroceptive and proprioceptive sensors. The perception layer utilizes the information obtained through the sensor layer for environment understanding \cite{Muresan}. The decision from the perception is fed to the planning layer that devises the optimal trajectories for the autonomous vehicle \cite{swn}. Finally, the control layer is responsible for the safe execution of control commands applied to the vehicle through lateral and longitudinal control \cite{Kim} \cite{Arkan} \cite{Azam}.
\par 
The primary goal is to understand the environment surrounding the autonomous vehicle through the fusion of exteroceptive and proprioceptive sensor modalities \cite{Munir2018}. The perception of the surrounding environment includes many challenging tasks, for instance, lane extraction, object detection, and traffic mark recognition, which provides the foundation for the safety of autonomous vehicles as standardized by the Safety of the Intended Functionality SOTIF-ISO/PAS-21448\footnote{https://www.daimler.com/innovation/case/autonomous/safety-first-for-automated-driving-2.htm}. The fundamental task in the hierarchy of perception is the extraction of lane information, as it assists an autonomous vehicle in precisely determining its position between the lanes. Accurate lane extraction forms the basis for the robust plans of autonomous vehicles, which includes lane departure and trajectory planning.
\par
In the literature, much promising research has been proposed based on either using handcrafted features or using an end-to-end deep neural network for lane detection using conventional frame-based RGB cameras \cite{Deusch2012} \cite{Jung2013} \cite{Kim2014} \cite{Li2016}. 
Conventional frame-based RGB camera performance is limited in various extreme and complex scenes \cite{Gallego2019}. For instance, by using conventional frame-based RGB cameras, the variation in illumination conditions can affect the performance of the lane detection algorithm because of unclear scenes in the input. Moreover, motion blur is typical for frame-based images when acquired from moving vehicles. The development of event cameras offers a promising solution to overcome uncertainty in conventional frame-based cameras caused by capturing the image at regular intervals. Event cameras capture per-pixel brightness changes, and each pixel streams the data asynchronously. Compared to the frame-based camera, the event camera provides a significant advantage in higher temporal resolution, high dynamic range and less motion blur.
\begin{figure*}[t]
      \centering
      \includegraphics[width=17cm]{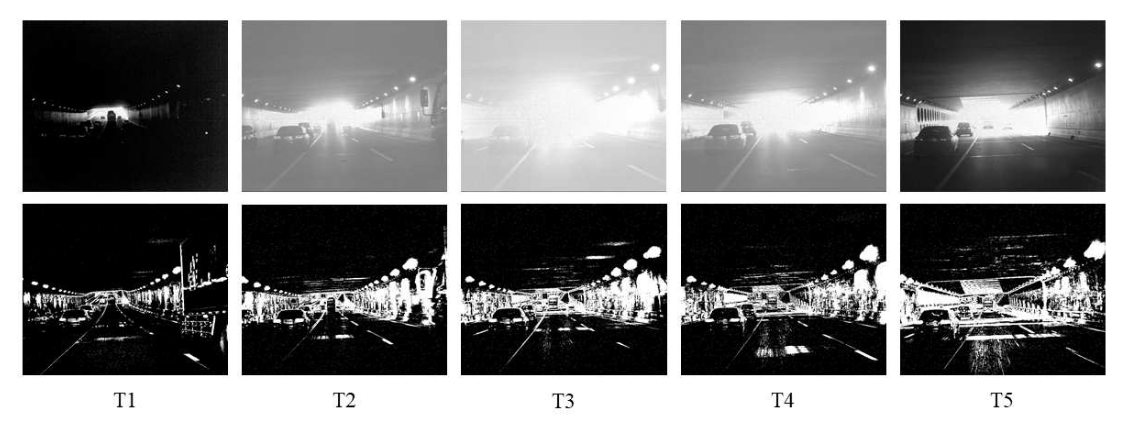}
\caption{A sequence of images captured while coming out of a tunnel (T1-T2-T3-T4-T5). The top row shows the grayscale images, and the bottom row shows the corresponding event camera images. RGB cameras are highly affected by illumination variations due to their low dynamic range. The figure is borrowed from \cite{Cheng2019} to illustrate the difference between event cameras and frame-based RGB cameras.}
      \label{tunnel}
\end{figure*}
Event cameras have two primary characteristics: i) a low latency rate and ii) a high dynamic range. An event camera captures the environment by the change in events, and its low latency rate helps generate the image faster than conventional frame-based cameras \cite{Gallego2019}. Additionally, this characteristic ensures that the image quality is not affected by motion blur. The high dynamic range of event cameras addresses the effect of illumination. Compared to conventional frame-based cameras having a dynamic range of $60$ dB, event cameras (for the DET dataset CeleX-V) provide a high dynamic range of $120$ dB that mitigates the illumination variation problem that appears in conventional frame-based cameras for lane detection \cite{Gallego2019} \cite{Mueggler2017}. Fig. \ref{tunnel} illustrates the difference between event cameras and standard conventional cameras.
\par 
 The perception of the environment plays an essential role in the architecture of the autonomous vehicle by determining the surrounding traffic entities, for instance, object detection. The inclusion of event cameras in the sensor suite of the autonomous vehicle provides an extra edge in the perception pipeline of the autonomous vehicle \cite{Chen}. Changes in illumination, sun glare, and motion blur are detrimental to frame-based cameras and lead to performance degradation of the perception module and may lead to autonomous vehicle fatalities. Moreover, in contrast to frame-based cameras, event cameras with low latency can benefit the perception module \cite{Gallego2019}. Notably, different exteroceptive sensors have pros and cons, but their integration them for in the autonomous vehicles provides a complement to different sensor modalities. This redundant integration of sensors contributes to the safety of autonomous vehicles in the environment.

\par
In this work, inspired by the utilization of event cameras in autonomous driving for lane detection tasks, as illustrated in \cite{Cheng2019}, an encoder-decoder neural architecture is designed for lane detection using an event camera. The architecture of the network is composed of three core blocks: i) an encoder, ii) an atrous spatial pyramid pooling (ASPP) block \cite{a3} \cite{a4}, and iii) an attention-guided decoder. The encoder of the proposed network is a combination of convolutional layers followed by a DropBlock layer and a max-pooling layer for the fast encoding of the input data. The ASPP block processes the encoded feature maps for the extraction of long-range features to ameliorate the spatial loss. The proposed decoder is based on an attention-guided decoder and is followed by fully connected layers to produce the lane detection predictions. The generality of the proposed network is experimentally validated on an event camera dataset. The event camera dataset contains various lane types, for instance, a single solid line, a single dashed line, a parallel solid line and a dashed line, and parallel dashed lines, etc. In addition, different numbers of lanes in the event dataset are collected by driving on roads with various carriageways. The event dataset also includes scene sparsity in addition to lane diversity. Therefore, the dataset is collected in various traffic scenes, for instance, driving on overpasses and bridges and in tunnels and urban areas, etc. Furthermore, in real-world scenarios, the viewpoint of the image plays an essential role in scene understanding. In this context, the event camera dataset is collected by changing the camera's location to increase the dataset's intraclass variance. The proposed 
lane marking detection network (LDNet) 
trained on this dataset handles complex scenes by incorporating the generalization of the scene sparsity and lane sparsity. The experimental evaluation of the proposed method is extensively tested on the event camera dataset, the dynamic vision sensor (DVS) dataset for lane extraction (DET), for multiclass and binary-class lane detection tasks and evaluated using the F-measure ($F1$ score) and intersection over union ($IoU$) metrics. The proposed method achieves a significant improvement of $5.54\%$ and $5.03\%$ on the mean $F1$ scores in the multiclass and binary-class tasks, respectively, surpassing the best-performing state-of-the-art method. In the case of the $IoU$ scores, the proposed method surpasses the best-performing state-of-the-art method by $6.50\%$ and $9.37\%$ in multiclass and binary-class tasks, respectively.
\par
Moreover, an ablation study is conducted on the Carla-DVS dataset and Event-Segmentation dataset. Carla-DVS is a synthetic dataset collected using the open-source Carla simulator.  The data consist of event data and binary labels for lane detection. The dataset is evaluated on the proposed algorithm and compared with other state-of-the-art algorithms. The LDNet is evaluated for generalization over the Event-Segmentation dataset.

\par
In summary, the main contributions of this work are as follows:
\begin{enumerate}

             \item   The novelty of this work is in the design of a convolutional encoder-decoder network for lane segmentation using the event camera dataset. We studied the encoder-decoder architecture for the lane detection task using the frame-based RGB camera as a sensor modality. We designed the encoder-decoder architecture for the novel application of lane detection using an event camera as the sensor modality based on the relevant literature. We present a detailed comparative analysis of our encoder-decoder framework and other state-of-the-art frameworks in Section II Related Work.
             \item  In this work, we have proposed a convolutional encoder-attention-guided decoder architecture in LDNet for lane marking detection using an event camera. The encoder architecture is composed of four convolutional layers followed by DropBlock layers to handle the event data lanes and scene sparsity. In the proposed method, the reason for using few convolutional layers is to ensure that the feature size computable (to avoid gradient explosion) because of the sparsity of event data. In addition, to retain the spatial resolution of the encoded features, a dense ASPP block is employed. The additive attention mechanism is utilized in the decoder part because of its better performance for high dimensional input encoded features that help improve lane localization and relieve postprocessing computation.
             \item In the proposed work, we employed the ASPP module to retain the spatial resolution by increasing the receptive field. The ASPP allows capturing valuable features as well as objects at multiple scales. The novelty of ASPP in the proposed work is the application to the event camera dataset for lane marking detection. Since using deeper convolutional neural networks (CNN) causes loss of spatial information at multiple scales and due to the sparsity of input data, it reduces the network performance. Furthermore, in contrast to DeepLabv3, we used the deep ASPP block for the feature extraction for lane marking detection using the event camera data. The deep ASPP block enables learning the feature representation at multiple scales and is followed by the attention-guided decoder module for lane marking detection. We have evaluated our method against DeepLabv3 and achieved obtained improvement of $15.82\%$ in the mean $F1$ score and $15.49\%$ in the mean $IoU$ score. 
\end{enumerate}
 This work addresses the novel problem of lane detection using an event camera by designing a convolutional encoder-attention-guided decoder architecture. The design of the encoder network consists of four convolutional blocks followed by DropBlock layers to address the lane and scene sparsity. In addition, to retain the spatial resolution of the encoded features, we have employed the deep ASPP module. Finally, we have added the attention-guided decoder that helps the proposed method better generalize for the lane detection task and relieve post-processing computations. The efficacy of the proposed method is extensively evaluated on the DET dataset and showing better performance in contrast to other state-of-the-art methods. The remainder of the paper is organized as follows: Section II introduces the event camera and its principle of operation related to the proposed method. Section III covers the related work. Section IV discusses the proposed methodology. Section V focuses on the experiments and results. The experimental analysis is discussed in Section VI. The ablation study is performed in Section VII and finally, Section VIII concludes the paper.

\section{Event Camera and Principle of Operation}
Event cameras are operated asynchronously in contrast to traditional frame-based cameras. The event camera captures the change in brightness (events) for each pixel independently in addition to capturing dense brightness, as in frame-based cameras at a fixed rate. In event cameras, the light is sampled by considering the scene dynamics with no dependencies related to the external clock (for instance, 30 fps (frame per second)) for the viewed scene. When measuring brightness changes, event cameras generate the sparse signals that are asynchronous in space and time, usually encoding moving image edges. This enables the event camera to obtain an advantage over traditional frame-based cameras in terms of high temporal resolution, low latency, low power computation and high dynamic range (($140$ dB and $120$ db) vs $60$ dB of standard cameras) \cite{Gallego2019} \cite{r2}.  
\par
The event camera generates the output in the form of events or spikes. The usability of these data to the convolutional neural network is to transform it into an apposite representation (for instance, images). In this context, the stream of event data is converted to an image where independent pixels correspond to a change in brightness, specifically, the logarithmic brightness signal $H(\mathbf{v_i},t_i)\doteq log I(\mathbf{v_i},t_i)$. For the pixel location $\mathbf{v_i}=(x_i,y_i)^T$ and time $t_i$, an event is recorded when the change in the brightness reaches the threshold ($\sigma$) \cite{r3}: 
\begin{equation}
\label{equs}
 H(\mathbf{v_i},t_i)-H(\mathbf{v_i},t_i-\triangle t_i)\geq p_i\sigma
\end{equation}
where $p_i \in {\{-1,1\}}$ corresponds to the polarity of brightness change and $\triangle t_i$ represents the time since the last event triggered at location $v_i$. A sequence of events $E(t_N)={e_i}_{i=1}^{N}={(x_i,y_i,t_i,p_i)}_{i=1}^{N}$ is generated in the time interval $\triangle t_i$. Eq.\ref{equs} represents the event generation model for the ideal sensor. In this work, we used the event dataset generated using the CeleX-V event camera \cite{r4}. In this camera, instead of polarity , a new event packet is introduced $E(t_N)={e_i}_{i=1}^{N}={(x_i,y_i,t_i,a_i)}_{i=1}^{N}$ where $``a``$ corresponds to an in-pixel time-stamp or pixel logarithmic gray level value. The CeleX-V encodes the events to the image representation by accumulating the event along the time interval $\triangle t_i$ that is set to $30 ms$ for this dataset.  
\par 
The event camera is a new sensor modality in contrast to frame-based traditional cameras, requiring the same maturity level of research as conducted on frame-based traditional cameras. The challenge in utilizing the event camera relies on processing event data, as agreement on the best method for representing the events has not yet been reached \cite{Gallego2019}\cite{r5}. The processing of event data is performed based on the application. As the event camera operates on the illumination variations in the scene, the utilization of the event camera inside a static scene will limit its usability. Notably, when the event camera is placed on a stationary vehicle with respect to the road and the scene dynamics are constantly changing, the camera generates data according to changes in brightness regardless of the stationary position of the vehicle.

\section{Related Work}
In autonomous driving, lane detection serves as a fundamental component, and much research has been focused on the development of robust lane detection algorithms \cite{Hillel2014}. In the literature, two types of mainstream techniques have been used for lane detection: traditional image processing methods and deep learning-based segmentation methods \cite{Aly2008} \cite{Wang2004} \cite{Huval2015}.
\par

\subsection{Traditional image processing methods for lane detection}
Traditional vision-based lane detection methods follow pipelines that include image preprocessing, feature extraction, lane model fitting and lane tracking. In traditional approaches, image preprocessing is a necessary step in determining the quality of features for lane detection tasks. For this purpose, image preprocessing includes region of interest (ROI) generation, image enhancement for extracting lane information and removal of non-lane information. The extraction of ROIs is an efficient method for reducing redundant information by selecting the lower part of the image \cite{Wu2014} \cite{Deng2013} \cite{Caceres2016}, and in some works, ROIs are generated using vanishing point detection techniques \cite{Son2015} \cite{Lee2018} \cite{Xu2016}. \cite{umar2017} has proposed a global way to estimate dense vanishing points using dynamic programming for multiple lane detection with horizontal and vertical curves .
Inverse perspective mapping (IPM) \cite{Du2016},\cite{Jung2015} or warp perspective mapping \cite{Shin2015} is used after ROI generation based on the parallel line assumption to reduce the effect of noise and to conveniently extract lanes. Lane enhancement is performed by using either color-based techniques or edge detection methods, such as hue-saturation-intensity (HSI) \cite{Sun2006}, luma, blue-difference and red-difference chroma components (YCbCr) \cite{Son2015}, and lightness, red/green and yellow/blue coordinates (LAB) \cite{Ma2010} as color-based models for transformation, and the Sobel operator \cite{De2015},\cite{Wang2012} and Canny detector \cite{Yoo2013} \cite{Niu2016} as edge-based techniques. Hybrid methods comprising color and edges are also used in research \cite{Caceres2016}. ROI generation reduces the noise in images, but it is not robust to shadows and vehicles. Filters are used in some works to eliminate non-lane information \cite{Nan2016} \cite{An2013} \cite{Wu2014}. In traditional approaches, lanes can also be modeled in the form of lines\cite{Liang2017} \cite{Nan2016}, parabolas\cite{S2015}\cite{Wang2012}, splines\cite{Nan2016}\cite{Wang2004} \cite{Kim2008}, hyperbolas\cite{Xu2016}, and so on. \cite{jiao2019} solved the lane detection problem by formulating it as a two-dimensional graph search problem. They designed a graph model that incorporates the continuous structure of lanes and roads. Furthermore, dynamic programming is used to solve the shortest path problem for the lane detection defined as the graph model.
Additionally, tracking is used as the postprocessing step to overcome illumination variations. Kalman filtering and particle filtering are the most widely used approaches for tracking lane detection \cite{Liang2017} \cite{Shin2015}. In addition to tracking, the authors also utilized Markov and conditional random fields as a postprocessing approach for lane detection \cite{Philipp2011}. \cite{Yuan2018} used a normal map for lane detection. The authors utilized the depth information for the generation of normal maps and used adaptive threshold segmentation for lane extraction.
\par
The traditional image processing methods for lane detection are unreliable for event camera data due to the nature of the data. The event camera data are sparse and consist of spikes, which indicate a change in brightness at each pixel. It lacks color information and complex information of scenes present in frame-based RGB images. The aforementioned techniques cannot be directly applied to event camera data, such as edge detectors, line fitting, Hough transforms, etc. They require human supervision and fail to extract valuable features that would affect the robustness of the lane detection task. 
\par
\subsection{Deep learning-based segmentation methods}
Recent advances in neural network architectures have exhibited a tremendous impact on refining the extracted features for lane detection tasks. The fine-tuning step of traditional methods in ROI generation, filtering and tracking has been solved by the use of neural networks. The deep neural networks formalize the lane detection problem as a semantic segmentation task. The vanishing point guided network (VPGNet) is guided by vanishing points for road and lane marking detection \cite{Lee2017}. \cite{Wang2018} proposed LaneNet, which performs detection in two stages: i) lane edge proposal generation and ii) lane localization. PolyLaneNet uses a front-facing camera for lane detection by generating the polynomials for each lane in the image via deep polynomial regression \cite{Tabelini2020}\cite{koy}. In \cite{Qin2020}, the authors formulated lane detection as a row-based selection problem using global features. The use of row-based selection has reduced the computational cost of lane detection tasks. Moreover, the self-attention distillation (SAD) approach is also used in lane detection tasks that allow model self-learning with any additional labels \cite{Hou2019}. \cite{Pizzati2019} used two cascaded neural networks in an end-to-end lane detection system. 
\cite{you2021} proposed a lane line detection technique. The network consists of two parts. First, a simple module follows an encoder-decoder architecture that learns features and predicts reasonable lanes. To handle more complex scenes, a second multitarget segmentation module is developed based on Wasserstein generative adversarial network (GAN). 

\par 
In the literature, the encoder-decoder architecture is broadly used for semantic segmentation using the image data. Lane marking detection, as the peculiar task of semantic segmentation, is used to classify the lanes in binary and multiclass categories. In the research domain, the most effective method to employ for lane marking detection is encoder-decoder architecture. In \cite{badrinarayanan2017segnet}, the authors designed  SegNet, an encoder-decoder architecture for image semantic segmentation. The encoder of the architecture consists of 13 convolutional layers inspired by the VGG-16 network followed by batch normalization, rectified linear unit (ReLU) activation function and a max-pooling layer. Each encoder has a corresponding decoding layer for upsampling the encoded feature map to full input resolution feature maps for semantic segmentation. The use of successive convolution results in spatial information loss, and due to sparsity of event data, in our proposed network, we use fewer convolution blocks with atrous spatial pooling to cater to information loss and improve lane detection.
    \cite{chougule2018efficient} proposed an encoder-decoder network for lane detection that is similar to SegNet \cite{badrinarayanan2017segnet} architecture. They removed the pooling layers and fully connected layers from the Segnet architecture and used the low-resolution image to achieve real-time lane detection. Furthermore, they trained the network by generating their dataset as a set of points for the lanes from the TuSimple dataset. The reduction in the resolution of the input image in the case of event data causes significant information loss, which is unfavorable for a robust lane detector.
\begin{figure*}[t]
      \centering
      \includegraphics[width=17cm]{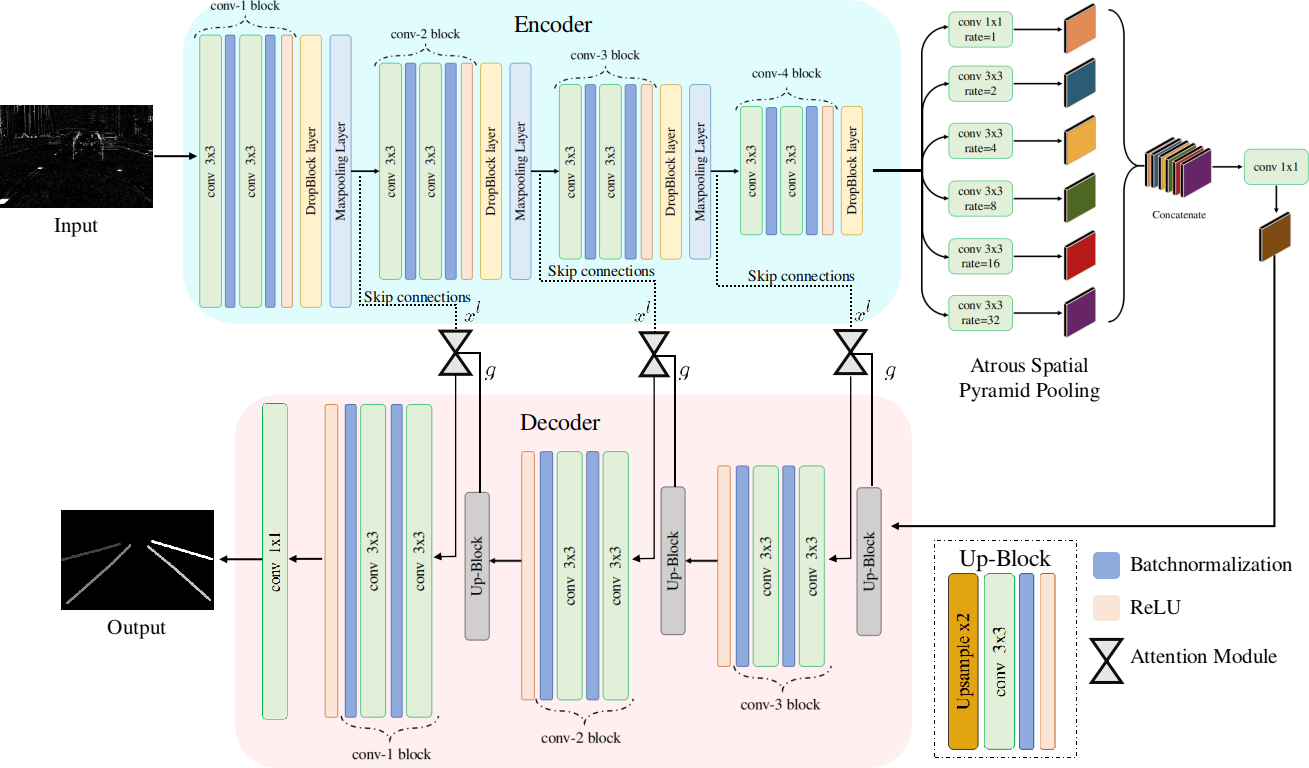}
\caption{The proposed LDNet architecture. The network is composed of three core components: i) a convolutional encoder to learn the feature representation, ii) a deep atrous spatial pyramid pooling (ASPP) block to retain the spatial resolution of encoded features, and iii) an attention-guided decoder. The attention-guided decoder is comprised of Up-block layers followed by the same convolutional block of the encoder part. The network takes an event camera image and predicts the lane marking detection.}
      \label{architecture}
\end{figure*}
\par

 In addition, some works have employed two decoders and semantic segmentation in an encoder-decoder architecture. For instance, \cite{ding2020lane} used the VGG16 network as a base model for the encoder followed by dilated convolution layers and two separate decoders for the semantic and instance segmentation for the lane detection task. In \cite{li2020efficient}, inspired by spatial pyramid pooling, the authors designed an encoder-decoder network. The main focus of their work is on designing the encoder network that includes an efficient dense module of depthwise separable convolution (EDD) and a dense spatial pyramid (DSP) module. For the decoder, they utilized bilinear interpolation and deconvolution for upsampling the encoder feature maps. \cite{sun2019accurate} designed an encoder-decoder network on the top of LaneNet architecture by replacing the LaneNet encoder with a sequential combination of atrous ResNet-101 and the spatial pyramid pooling (SPP) network. The decoder module consists of two decoder networks for embedding the feature map and binary segmentation map.
\par 
This work employed the encoder-attention-guided decoder architecture for lane marking detection using event camera data. In contrast to the abovementioned encoder-decoder architecture, we used the DropBlock layer with convolutional blocks to retain the network generality for the lane detection task. In the SegNet \cite{badrinarayanan2017segnet} architecture and \cite{chougule2018efficient}, the spatial resolution is lost in the succession of the encoder architecture. We employed the ASPP module to retain the spatial resolution followed by the attention-guided decoder, which improves the localization of lanes. Moreover, in our work, the attention-guided decoder is beneficial to lane marking detection by not performing additional postprocessing steps and in contrast to \cite{ding2020lane}\cite{li2020efficient}\cite{sun2019accurate} optimizes the localization of features in the feature map.

\par
\subsection{Datasets for lane marking detection}
The most common datasets used by the traditional and deep learning-based methods include the Caltech dataset \cite{Aly2008}, TuSimple dataset \cite{tusimple}, and CULane dataset \cite{Pan2017}. These datasets are based on RGB images generated by conventional cameras. The change in illumination and motion blur in the images will affect the performance of the lane detection algorithm. Event cameras are a type of novel sensor that address the problem of standard cameras by having a dynamic range and low latency. In the literature, several event camera datasets have been published, including the Synthesized Dataset \cite{Mueggler2017}, Classification Dataset \cite{Li2017}, Recognition Dataset \cite{Hu2016}, and Driving Dataset \cite{Binas2017}. The aforementioned event camera datasets are for general purposes, and none of them are explicitly dedicated to the lane detection task. Additionally, these datasets have low spatial resolutions. The two main applications that have been published in the research on the event camera include steering angle prediction \cite{24} and car detection \cite{Chen2018}. \cite{Cheng2019} proposed an event camera dataset for lane detection tasks. The authors evaluated their dataset with different lane detection algorithms, including DeepLabv3 \cite{a4}, a fully convolutional network (FCN) \cite{Long2015}, RefineNet \cite{Lin2017}, LaneNet \cite{Wang2018} and a spatial convolution neural network (SCNN)\cite{Pan2017}, and published the benchmark for lane detection tasks using event cameras. In their lane detection benchmark, the SCNN \cite{Pan2017} outperformed all the other algorithms and achieved a better mean $IoU$ and mean $F1$ score. Inspired by \cite{Cheng2019}, in this work, we use their dataset in the lane detection task, and the experimental evaluation of the proposed method surpasses the abovementioned benchmark in terms of both the mean $F1$ and $IoU$ scores.

\section{Methods}
In this section, we describe in detail the proposed framework for lane marking detection, as illustrated in Fig. \ref{architecture}. The framework consists of three modules: a convolutional encoder module, which extracts the features from the input image; a deep ASPP block to extract global features, and an attention-guided decoder module. In addition, skip connections are added from the encoder to the decoder to retain high-frequency spatial features.

\subsection{Encoder}
The CNNs outperform traditional computer vision and image processing techniques that incorporate handcrafted features for lane marking detection using standard RGB cameras \cite{Wu2014} \cite{Deng2013} \cite{Caceres2016}. However, lane marking detection with event cameras is a new research domain, and many state-of-the-art deep learning-based lane detection algorithms, such as SCNN \cite{Pan2017}, LaneNet \cite{Wang2018}, and FCN \cite{Long2015}, are implemented on event camera images but require further improvements in robustness \cite{Cheng2019}. 
 Fig.\ref{architecture} shows the design of the encoder for the event camera for lane marking detection. The related work gives an insight into the motivation of the designed encoder. We adopted and modified the encoder from the SegNet architecture \cite{badrinarayanan2017segnet}. SegNet  uses convolution blocks similar to the VGG architecture \cite{vgg}. In contrast to SegNet, the LDNet encoder has four operation blocks so that the feature map size is sufficient for ASPP to extract features, and on the other hand, it reduces the number of parameters in the LDNet encoder from 14.7M to 583.07 k.  Each block consists of a convolution stack,  a max-pooling layer and an additional DropBlock layer that improves the regularization of the LDNet. However, the last operational block does not include a max-pooling layer to match the filter size of the decoder. The encoder details of each layer are given in Table \ref{layer}.
\par
Since the convolution stack of the encoder architecture is adopted from the VGG architecture \cite{vgg}, the convolutional layer parameters in terms of the receptive filter size and stride are kept the same as those of the VGG architecture: $3\times3$ and $1$, respectively. To increase the detailed representation of low-level feature encoding, the convolution stack consists of two convolutional layers followed by batch normalization. A nonlinear activation function is employed after the second convolutional layer, which makes the decision function discriminative. Let $x^l$ be the higher-dimensional image representation extracted from convolutional layers by progressively processing local features layer by layer. This process categorizes pixels in higher-dimensional space corresponding to their semantics. However, the model predictions are conditioned on the features extracted from the receptive field. For each convolutional layer $l$, a feature map $x^l$ is obtained by sequentially applying a linear transformation realized by a nonlinear activation function. The ReLU function is chosen as a nonlinear activation function, as shown in Eq.\ref{1}:


\begin{equation}
 \sigma(x^l_{i,c})= max(0,x^l_{i,c}),
\label{1}
\end{equation}
where $c$ represents the channel dimension, $i$ denotes the spatial channel dimensions and $\sigma$ corresponds to the activation function. Eq.\ref{2} represents the feature map activation formulation. \\
\begin{equation}
\vspace{0.1cm}
\begin{aligned}
& x^l_{c}=\sigma_1( \sum_{{c'}\in F_l} x_{c'}^{l-1}*k_{c',c}),
\label{2}
\end{aligned}
\end{equation}
$*$ represents the convolution operation, $F_l$ is the number of feature maps in layer $l$, and $k$ is the convolution kernel. The subscript $i$ is ignored for notational clarity in the equation. The function $f(x^l;\Phi ^l)=x^{l+1}$ is applied to convolutional layer $l$, where $\phi ^l$ is a trainable kernel parameter. These parameters are learned by minimizing the objective function during training.
\par
The DropBlock layer is introduced after each convolution stack in the operational block, as inspired by \cite{DB}. It is a structured form of dropout that is particularly efficient in regularizing the CNN. The notable difference between DropBlock and dropout is that DropBlock drops the contiguous regions from a feature map rather than random independent values. The pseudocode of DropBlock is illustrated as Algorithm 1. $BS$ and $\gamma$ are the two main tuning parameters. The $BS$ represents the size of the block to be dropped, while $\gamma$ is a control parameter for the number of activation connections to be dropped. DropBlock is not applied during evaluations, similar to dropout. A max-pooling layer is incorporated in each operational block to reduce the size of the feature map.
\begin{table}[t]
\caption{The detailed architecture of the proposed LDNet.}
\label{layer}
\resizebox{8.5cm}{!}{%
\begin{tabular}{@{}l|c|c@{}}
\toprule
                                               & Layer                  & Output size \\ \midrule
                                               & Input data             &             \\ \midrule
\multirow{8}{*}{Encoder}                       & Conv-1 block           & 32*256*256  \\
                                               & DropBlock +Max pooling  & 32*128*128  \\
                                               & Conv-2 block           & 64*128*128  \\
                                               & DropBlock + Max pooling & 64*64*64    \\
                                               & Conv-3 block           & 128*64*64   \\
                                               & DropBlock + Max pooling & 128*32*32   \\
                                               & Conv-4 block           & 256*32*32   \\
                                               & DropBlock              & 256*32*32   \\ \midrule
\multirow{7}{*}{Astrous spatial Pooling block} & Conv 1X1 (rate=1)      & 256*32*32   \\
                                               & Conv 1x1 (rate=2)      & 256*32*32   \\
                                               & Conv 1x1 (rate=4)      & 256*32*32   \\
                                               & Conv 1x1 (rate=8)      & 256*32*32   \\
                                               & Conv 1x1 (rate=16)     & 256*32*32   \\
                                               & Conv 1x1 (rate=32)     & 256*32*32   \\
                                               & Concatenate + Conv 1X1 & 256*32*32   \\ \midrule
\multirow{9}{*}{Decoder}                       & Up-Block               & 128*64*64   \\
                                               & Attention Module       & 256*64*64   \\
                                               & Conv-3 block           & 128*64*64   \\
                                               & Up-Block               & 64*128*128  \\
                                               & Attention Module       & 128*128*128 \\
                                               & Conv-2 block           & 64*128*128  \\
                                               & Up-Block               & 32*256*256  \\
                                               & Attention Module       & 64*256*256  \\
                                               & Conv-1 Block           & 32*256*256  \\ \midrule
Output                                         & Conv 1X1               & 5*256*256   \\ \bottomrule
\end{tabular}%
}
\end{table}

\begin{algorithm}
\SetAlgoLined
\KwInput{feature map obtained from convolutional layer $x^l$, $BS$, $\gamma$, $mode$ }

\If{$mode==evaluation$}{
return $x^l$}
Randomly generate mask $M:M_{i,j}~ Bernoulli(\gamma)$\\
For each zero position $M_{i,j}$, a spatial square mask is created with size equal to $BS$ and centered at $M_{i,j}$\\
Set all the values inside the spatial square mask equal to zero\\
Apply the mask $x^l=x^l\times M$
Normalize the feature map $x^l=x^l \times count(M)/count-ones(M)$
\caption{DropBlock layer.}
\end{algorithm}

\subsection{Atrous spatial pyramid pooling block}
In CNNs, reducing the receptive field size results in the loss of spatial information, which is associated with repeated usage of max pooling and strided convolution completed in the successive layers. One possible way to decrease the spatial loss is the addition of deconvolutional layers \cite{Long2015} \cite{a2}, but it is computationally intensive. The notion of atrous convolution was introduced by \cite{a3} \cite{a4} to overcome the spatial loss problem. The dilated convolution operation increases the receptive field without increasing the training parameters or feature map resolution. 
Table \ref{layer} gives the details of the ASPP module. Here, we used a deeper ASPP module than those in \cite{a3} \cite{a4}, which helps LDNet in learning higher-dimensional features across the entire feature map and refining full-resolution lane marking detection for event data. The event data are sparse in nature, and reducing the feature map in successive convolution causes information loss. However, we used a deep ASPP module that helps to extract deeper features without reducing the feature map size.
\par
The atrous convolution operation is employed for one-dimensional or two-dimensional input data. Considering one-dimensional input data first, an atrous convolution is formalized as the output $y[i]$ of the input signal $x[i]$ with a kernel filter $w[k]$ of length $K$, as shown in Eq.\ref{3}:
\begin{equation}
y[i]=\sum_{k=1}^{K} x[i+r.k]w[k],
\label{3}
\end{equation}
where $r$ is the rate parameter that corresponds to the stride through which the input signal is sampled. The standard convolution is an atrous convolution with a rate of $r=1$. The variable $i$ represents the location on the output signal $y[i]$ when the atrous convolution having kernal filter $w[k]$ is applied on the input image $x[i]$. $k$ represents the indices of the atrous convolution kernel. The increase in the rate parameter increases the receptive field of the feature map at any convolutional layer without the increase in computation power and number of parameters. It introduces $r-1$ zeros in the consecutive filter values in the feature map, efficiently increasing the kernel size of the $K \times K$ filter to $K_d=k+(k-1)(r-1)$ without increasing the number of parameters or increasing the computational complexity. Therefore, it offers an effective mechanism to control the receptive field of view and find the best compromise between the localization of an object of interest and context assimilation. In this work, the feature enhancer module consists of a deep ASPP block. The feature map obtained from the encoder module is convolved with the deep ASPP block. It consists of six layers with a rate ranging from $r=2^0$ to $2^5$. The output from each layer is concatenated and given to the attention-guided decoder block.
\subsection{Attention-guided decoder}
The semantic contextual information is captured efficiently by the acquisition of a large receptive field, and for this step, the feature map is gradually downsampled in a typical CNN. The features on the coarse spatial grid model the location and their relationship with different features at the global level. However, reducing false-positive predictions for small objects with large variability is a challenging task. In this work, we propose a novel attention-guided decoder. 
Generally, in the literature, attention (additive \cite{bahdanau2014neural} and multiplicative \cite{luong2015effective}) and self-attention are used. In the proposed method, we have used additive attention \cite{bahdanau2014neural} to transfer the information from the encoder to the decoder. Choosing this attention in the proposed method enhances the feature representation and localization that progressively reduces the feature response in unrelated background regions without extracting the ROI. In this attention mechanism, the decoder neurons receive the additional input through attention from encoder states/activation providing more flexibility in terms of what to focus on from a regional basis when combined with gating signal from the coarsest scale activation map. In addition, the use of additive attention in comparison to multiplicative attention is employed because the additive attention tends to perform better for high dimensional input features.
\par
We have not employed the self-attention mechanism in the proposed method because, in the self-attention mechanism, only the attention is applied within one component. The objective of the proposed method is to detect the lane marking, and for this task, an encoder-decoder architecture with the addition of the ASPP module as a spatial feature enhancer is employed. In the proposed method, if self-attention is employed, then the decoders usability is limited, or no decoder will be used, as in bidirectional encoder representations from transformers (BERT) \cite{devlin2018bert} architecture.
\par
The attention coefficient $\alpha_i \in[0,1]$ in the attention-guided decoder distinguishes prominent image regions and prunes features from task-specific activations. The output of the attention module is the elementwise multiplication of attention coefficients and input feature maps described in Eq. \ref{4}:
\begin{equation}
\vspace{0.1cm}
\begin{aligned}
 & \hat{x}^l_{i,c}=x^l_{i,c}\cdot \alpha^l_i,
 \label{4} 
\end{aligned}
\end{equation}

\begin{figure}[t]
\centering
\includegraphics[width=8.5cm]{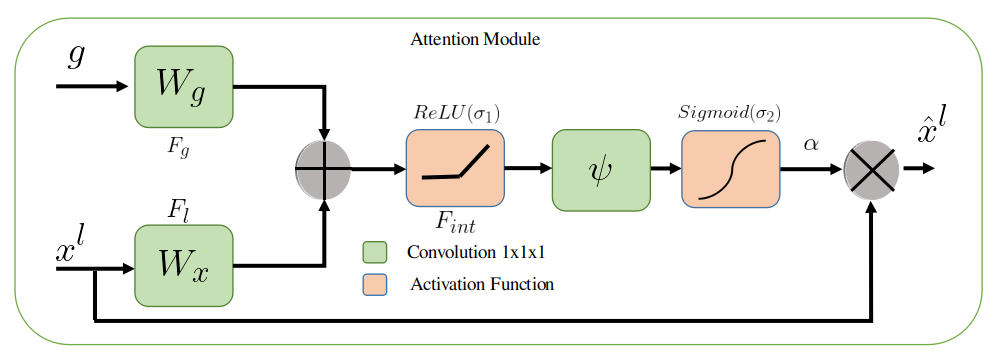}
\caption{The working operation of the attention-guided decoder is illustrated. The term $g$ corresponds to the vector taken from the lowest layer of the decoder. In our case, it is taken from the Up-Block layer. $x^l$ represent the encoded features from the encoder network. }
\label{atten}
\end{figure}
For each pixel vector $x^l_i \in \mathbb{R}^{F_l}$, a single scalar attention value is calculated. \cite{a5} proposed a multidimensional attention coefficient to learn sentence embeddings. Since lane marking detection is a multiclass problem, we utilize multidimensional attention coefficients to learn the semantic context in the image. Fig. \ref{atten} shows the attention module. The input vector $g_i \in\mathbb{R}^{F_g}$ determines the focus region for each pixel $i$. Eq.\ref{5} shows the additive attention formulation.
\begin{figure*}[t]
      \centering
      \includegraphics[width=17cm]{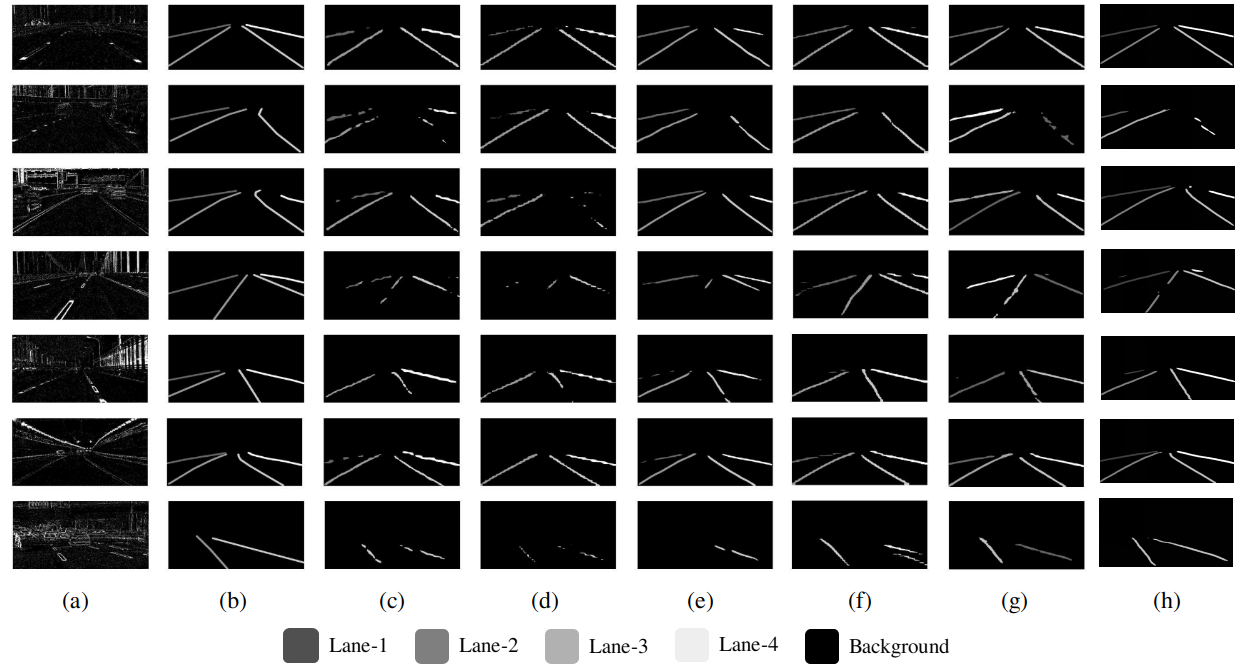}
\caption{The qualitative comparison between different lane detection methods using multiclass labels. There are 5 classes: background, lane-1, lane-2, lane-3 and lane-4.} (a) shows the input image. (b) shows the ground-truth labels. (c-h) show the results for FCN, DeepLabv3, RefineNet, SCNN, LaneNet and LDNet (ours), respectively. 
      \label{multi-class}
\end{figure*}

\begin{equation}
\begin{aligned}
 & q^l_{att}=\psi^T (\sigma (W^T_xx^l_i+W^T_gg_i+b_g))+b_\psi, \\
 & \alpha ^l_i=\sigma _2(q^l_{att}(x^l_i,g_i;\Theta_{att} )), 
 \label{5}
\end{aligned}
\end{equation}
Here, $\sigma _2(x_{i,c})=\frac{1}{1+exp(-x_{i,c})}$ represents the sigmoid activation function. $\Theta_{att} $ characterizes a set of parameters including the linear transformation $W_x \in \mathbb{R}^{F_l \times F_{int}}$, $W_g \in \mathbb{R}^{F_g \times F_{int}}$, $\psi  \in \mathbb{R}^{F_{int} \times 1}$ and bias term $b_\psi \in \mathbb{R}$, $b_g \in \mathbb{R}^{F_{int}}$.  The term $g$ represent the vector taken from the lowest layer of the network. The channel-wise $1 \times 1 \times 1$ convolutions compute the linear transformation for the input tensors, which is called ``vector concatenation-based attention`` and involves concatenating the features $x^l$ and $g$ and linearly mapping to a $\mathbb{R}^{F{int}}$ multidimensional space \cite{a6}. There are three operational blocks in the attention-guided decoder. Each block consists of a convolutional stack that is similar to the encoder, an Up-Block layer to increase the feature map size, and the attention module. The Up-Block layer includes an upsampling layer followed by convolutional, ReLU and batch normalization layers. The attention module highlights the salient features that are carried through the skip connections, as shown in Fig \ref{architecture}. The features obtained at the coarse scale are used in gating to remove irrelevant and noisy skip connections. 
Gating is performed 
before concatenation to add only relevant activations, as in Fig. \ref{atten}. A fully connected layer is added following the decoder module, which classifies each pixel in the feature map and is further compared with the corresponding ground truth to calculate the loss during training.
\par 
 The proposed encoder-decoder network is jointly trained in end-to-end manner for lane marking detection. During the training, the loss is backpropagated to optimize the weight of the network. We incorporated the cross-entropy function given by  Eq.\ref{loss}:
\begin{equation}
\begin{aligned}
Loss=-\frac{1}{N}\sum_{N}^{j=1}\sum_{M}^{c=1}y_{c,j} \ln (\hat{y}_{c,j})
 \label{loss}
\end{aligned}
\end{equation}
where $N$ is number of pixels in the ground truth and $M$ is the number of classes. $y_{c,j}$ defines the ground truth of a pixel belonging to the correct class, and $\hat{y}_{c,j}$ defines whether the predicted pixel belongs to correct class $c$. 
\section{Experiments and Results}
The effectiveness of the proposed method for lane marking detection in event camera-based images (DET dataset \cite{Cheng2019}) is evaluated using multiclass and binary-class labels. The results are compared with the state-of-the-art algorithm benchmark on the DET dataset. The proposed method is evaluated in terms of the mean $F1$ score and the mean $IoU$. The details are described below.
\subsection{DET dataset}
In our experiments, we use the benchmark developed by \cite{Cheng2019}.
A high-resolution dynamic vision sensor dataset for lane detection is published. The dynamic vision sensor is a type of event-based sensor that responds to variations in brightness. It does not follow the principle of frame-based RGB cameras, 
but individual pixels are incorporated in the sensor function individually and asynchronously, recording variations in brightness. The DET dataset is collected using a CeleX-V event camera with a resolution of $1280 \times 800$ mounted on a car. The dataset is recorded at different times of day and comprises various traffic scenes, such as urban roads, tunnels, bridges, and overpasses. The dataset also includes various lane types, such as parallel dashed lines, single lines, and single dashed lines.
The DET dataset consists of a total of $5424$ images with binary and multiclass labels. In the case of multiple classes, the labels are categorized into five classes, where four labels correspond to different lane types and the last label is for the background. In this work, we use both (lanes and background) types of labels to evaluate the proposed method. For the experimental evaluation, the dataset is split into training, validation and test data at percentages of 50\%, 16\%, and 33\%, respectively.


\subsection{Training details}
The proposed LDNet is implemented using the PyTorch deep learning library. The network is trained from scratch in an end-to-end manner. A $256 \times 256$ size image is input to the network. The input images does not have any preprocessing or filtering step in our network. The network learns to filter out white noise present in the images.  
The Adam optimizer is adopted for training the network with the learning rate schedule policy given by Eq.\ref{6}. The initial learning rate of $5^{-4}$; epsilon is set to $1^{-8}$, and the weight decay is set to $1^{-4}$. In Eq.\ref{6}, the value of $power$ in training the network is set to $0.9$. The tuning of neural network parameters involves heuristics, or some parameters are architecture-specific, but in the literature, 
some parameters are considered to have been perfected after years of studies
. In this work, we conducted extensive experimentation and evaluation to determine the best parameters for the proposed method. In all of our experiments, we kept the same parameter settings.
\begin{equation}
LR=initial-LR \times (\frac{1-epoch}{max-epochs})^{power},
\label{6}
\end{equation} 
\par
In addition, the DropBlock parameters mentioned in Algorithm. 1, i.e., $BS$ and $\gamma$, are also fixed when training the proposed network. The value of $BS$ is fixed at $5$, whereas the $\gamma$ value is determined by Eq.\ref{7} for controlling the features to be dropped during training.

\begin{equation}
\gamma =\frac{1-kP}{BS^2}\frac{feat^2}{(feat-BS+1)^2},
\label{7}
\end{equation}
where $kP$ defines the probability of keeping a unit. In our experiments, the value of $kP$ is linearly increased from $0.0$ to $0.5$. $feat$ denotes the feature map size. These hyperparameter values are inspired by \cite{DB} and were selected empirically by using grid search. 

The proposed method is evaluated with both label categories, i.e., multi-class labels and binary class labels. However, in experimenting with both labels, the training parameters of the proposed network are kept the same. The training process runs for a total of $100$ epochs, with batch size of $4$ using PyTorch deep learning library on an Nvidia RTX 2060 GPU.

\subsection{Evaluation metrics}
\label{marker}
The research society has matured and standardized the evaluation metric for lane marking detection. The public frame-based lane detection benchmarks have utilized $F1$ and $IoU$ scores to evaluate lane marking detection \cite{2} \cite{3}. Moreover, in the literature, to evaluate event camera segmentation \cite{EVnet} and lane marking detection \cite{Cheng2019}, $F1$ and $IoU$ scores are adopted. Notably, in evaluating the proposed LDNet, the image size is kept at $256 \times 256$. In this work, we have also used the mean $F1$ and $IoU$ scores to evaluate the proposed method. The $F1$ score is expressed in Eq.\ref{8}: \\
\begin{equation}
F1=2 \times\frac{P\times R}{P + R},\\
\label{8}
\end{equation}
\begin{equation}
 P=\frac{TP}{TP+FP},\\
\label{9}
\end{equation}
\begin{equation}
R=\frac{TP}{TP+FN},\\
\label{10}
\end{equation}
where $TP$, $FP$ and $FN$ represent the number of true positives, false positives, and false negatives, respectively. The $IoU$ is given by Eq. \ref{11}: \\
\begin{equation}
 IoU(S_m,S_{gt})=\frac{\mathbb{N}(S_m\bigcap S_{gt})}{\mathbb{N}(S_m\bigcup S_{gt})},
\label{11}
\end{equation}
where $S_m$ represents the predicted lane detection output and $S_{gt}$ denotes the ground-truth labels. $\bigcap$, $\bigcup$, and $\mathbb{N}$ represent
intersection, union and number of pixels, respectively. We evaluated the mean $F1$ and $IoU$ scores for both multiclass and binary-class lane detection.

\begin{table}[t]
\centering
\caption{Comparison of evaluation results of LDNet with other state-of-the-art methods on the DET dataset. The mean $F1$ scores ($\%$) and mean $IoU$s ($\%$) are used as evaluation metrics for the multiclass labels. The values in bold are the best scores.}
\label{table 1}
\resizebox{8.5cm}{!}{%
\begin{tabular}{@{}ccc@{}}
\toprule
Model                                    & Mean $F1$ (\%)                    & Mean $IoU$ (\%) \\ \midrule
\multicolumn{1}{c|}{FCN \cite{Cheng2019}}       & \multicolumn{1}{c|}{60.39} & 47.36    \\ \midrule
\multicolumn{1}{c|}{DeepLabv3 \cite{Cheng2019}} & \multicolumn{1}{c|}{59.76} & 47.30    \\ \midrule
\multicolumn{1}{c|}{RefineNet \cite{Cheng2019}} & \multicolumn{1}{c|}{63.52} & 50.29    \\ \midrule
\multicolumn{1}{c|}{LaneNet \cite{Cheng2019}}   & \multicolumn{1}{c|}{69.79} & 53.59    \\ \midrule
\multicolumn{1}{c|}{SCNN \cite{Cheng2019}}      & \multicolumn{1}{c|}{70.04} & 56.29    \\ \midrule
\multicolumn{1}{c|}{\textbf{LDNet-multiclass (ours)}}  & \multicolumn{1}{c|}{\textbf{75.58}} & \textbf{62.79} \\ \midrule
\end{tabular}%
}
\end{table}
\begin{table}[]
\centering
\caption{Comparison of the evaluation results of LDNet with other state-of-the-art methods on the DET dataset. The mean $F1$ scores ($\%$) and mean $IoU$s ($\%$) are used as evaluation metrics for the binary class labels. The values in bold are the best scores.}
\label{table 2}
\resizebox{8.5cm}{!}{%
\begin{tabular}{@{}ccc@{}}
\toprule
Model                                    & Mean $F1$ (\%)                    & Mean $IoU$ (\%) \\ \midrule
\multicolumn{1}{c|}{FCN}       & \multicolumn{1}{c|}{72.65} & 58.51    \\ \midrule
\multicolumn{1}{c|}{DeepLabv3 } & \multicolumn{1}{c|}{71.93} & 58.45    \\ \midrule
\multicolumn{1}{c|}{RefineNet} & \multicolumn{1}{c|}{75.78} & 61.44    \\ \midrule
\multicolumn{1}{c|}{LaneNet }   & \multicolumn{1}{c|}{79.21} & 64.74    \\ \midrule
\multicolumn{1}{c|}{SCNN }      & \multicolumn{1}{c|}{80.15} & 67.34    \\ \midrule
\multicolumn{1}{c|}{\textbf{LDNet-binary class (ours)}}  & \multicolumn{1}{c|}{\textbf{85.18}} & \textbf{76.71} \\ \midrule
\end{tabular}%
}
\end{table}

\subsection{Results}
The DET dataset is benchmarked on typical lane detection algorithms, which include the FCN \cite{Long2015}, RefineNet \cite{Lin2017}, SCNN \cite{Pan2017}, DeepLabv3 \cite{a4} and LaneNet \cite{Wang2018} algorithms. The FCN algorithm is one of the earliest works to perform semantic segmentation by classifying every pixel in an image. An end-to-end FCN is trained to predict the segmentation map. DeepLabv3 investigates ASPP by upsampling a feature map to extract dense and global features. RefineNet explores a multipath refinement network that extracts features along the downsampling process to allow high-resolution predictions using long residual connections.
\par
However, LaneNet and SCNN were specifically designed for lane detection tasks. SCNNs achieve state-of-the-art accuracy on the TuSimple dataset \cite{tusimple}. They use slice-by-slice convolutions within feature maps to enable message crossing between pixels across rows and columns. LaneNet applies a learned perspective transformation trained on the images. For each predicted lane, a third-degree polynomial is fitted, and lanes are reprojected onto the images.
\par 
The aforementioned methods are considered baseline methods and compared with the proposed network. Table \ref{table 1} and Table \ref{table 2} show the evaluation of the proposed method to the baseline methods. LaneNet and SCNN outperform typical semantic segmentation algorithms such as FCN, DeepLabv3 and RefineNet. However, LDNet (the proposed method) outperforms the best-performing state-of-the-art SCNN with an improvement of $5.54 \%$ on the mean $F1$ score and $6.5\%$ on the mean $IoU$ for multiclass lane detection, and an improvement of $5.03 \%$ on the mean $F1$ score and $9.37\%$ on the mean $IoU$ for binary-class lane detection. This comparison provides insight into how the use of the ASPP module with an attention-guided decoder improves the detection of lane markings. It should be noted that no postprocessing step is utilized in our framework. Fig. \ref{multi-class} shows the qualitative results of the proposed algorithm with the baseline methods in multiclass lane detection.
\par 
 We calculated the FLOPs (floating point operations per second) and number of parameters for the proposed method and the state-of-the-art methods. Table \ref{FLOPS} illustrates the computational cost in FLOPs and the number of parameters.  The proposed network has $5.71M$ parameters and $12.49$ GMac\footnote{MACS is the abbreviation for the number of fixed-point multiply-accumulate operations performed per second. It is a measure of the fixed-point processing capacity of a computer. This amount is often used in scientific operations that require a large number of fixed-point multiply-accumulate operations.
A GMACS: equal to 1 billion ($=10 ^ 9$) fixed-point multiply-accumulate operations per second} FLOPs, second-best compared to other state-of-the-art algorithms. As the proposed LDNet has utilized the dense ASPP (the initial variant introduced by DeepLabV3) in an encoder-attention-guided decoder architecture, the proposed model has a lower computational cost and higher accuracy than DeepLabv3. 
\begin{table}[h]
\centering
\caption{Comparison of the computational costs of the proposed LDNet and other state-of-the-art methods in terms of FLOPs and number of parameters.}
\label{FLOPS}
\resizebox{9cm}{!}{%
\begin{tabular}{@{}ccc@{}}
\toprule
Model                                    & Number of parameters                    & FLOPS \\ \midrule
\multicolumn{1}{c|}{FCN}       & \multicolumn{1}{c|}{132.27 M} & 62.79 GMac    \\ \midrule
\multicolumn{1}{c|}{DeepLabv3 } & \multicolumn{1}{c|}{39.05 M} & 30.91 GMac    \\ \midrule
\multicolumn{1}{c|}{RefineNet} & \multicolumn{1}{c|}{99.02 M} & 46.51 GMac  \\ \midrule
\multicolumn{1}{c|}{LaneNet }   & \multicolumn{1}{c|}{0.526 M} & 0.64 GMac  \\ \midrule
\multicolumn{1}{c|}{SCNN }      & \multicolumn{1}{c|}{25.16 M} & 90.98 GMac   \\ \midrule
\multicolumn{1}{c|}{LDNet-binary class (ours)}  & \multicolumn{1}{c|}{5.71 M} & 12.49 GMac \\ \midrule
\end{tabular}%
}
\end{table}
\begin{figure}[t]
      \centering
      \includegraphics[width=9cm]{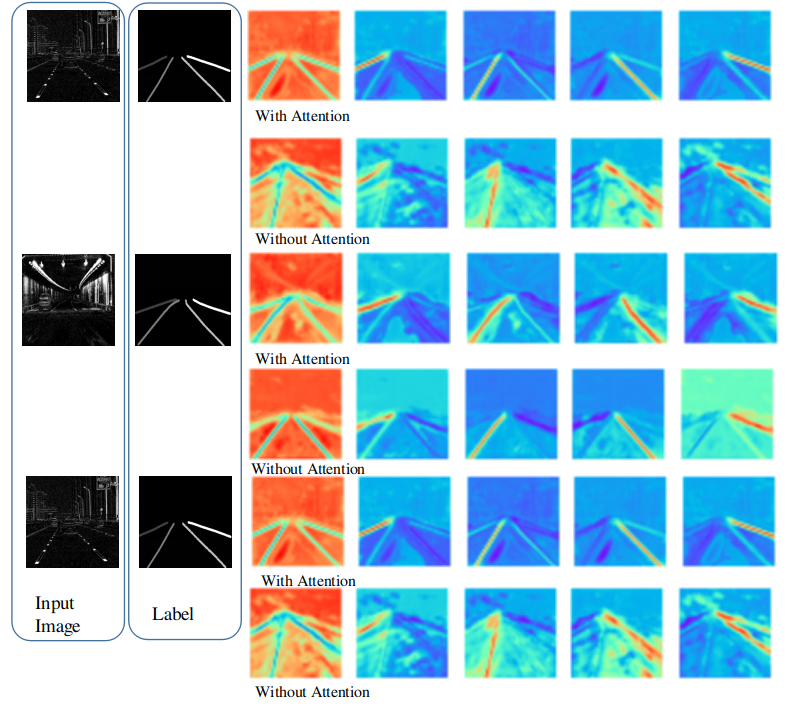}
\caption{The visualization of feature activation with and without the attention module in the decoder. The input image and the corresponding labels are also shown. The orange color shows the predicted class, and blue is the nonpredicted class. The first row from left shows the predicted background, lane 1, lane 2, lane 3 and lane 4. }
      \label{atten-map}
\end{figure}
\section{Experimental Analysis}
In this section, we investigate the effect of the different factors (using a backbone network before the encoder network, the addition of the DropBlock layer and the attention-guided decoder) on the performance of the proposed method.
\par
We experiment with the proposed network with a deeper encoder by utilizing six different backbone networks: VGG16 \cite{vgg}, ResNet-18 \cite{resnet}, ResNet-50 \cite{resnet}, MobileNetV2 \cite{mobilenet}, ShuffleNet \cite{shufflenet} and DenseNet \cite{densenet}. The image is fed to the backbone network, and the feature map is given to the proposed encoder. The pretraining weights are used for the backbone networks. Table \ref{table 3} shows the results when using the deeper encoder in the proposed network. The evaluation results show no significant gain from incorporating the backbone network compared to the proposed network. This finding justifies the use of shallow encoders in LDNet.
\par
Table \ref{table 4} shows the evaluation of the proposed network with DropBlock, spatial dropout, and no dropout. The dropout layer is added to the network to regularize the network and to prevent overfitting. The addition of the DropBlock shows improved results on the test dataset compared to no dropout or spatial dropout. The contiguous regions in the feature map are highly correlated; dropping random units still allows information flow but is not efficient in regularizing the network. The DropBlock helps the network retain semantic information required for lane marking detection.
\par
\begin{table}[t]
\centering
\caption{Quantitative analysis of LDNet with different backbone networks. The experimental analysis is performed on both the multiclass and binary-class tasks and the results are evaluated in terms of the mean $F1$ and $IoU$ scores.}
\label{table 3}
\resizebox{8cm}{!}{%
\begin{tabular}{@{}c|c|c|c|c@{}}
\toprule
\multirow{2}{*}{Model} & \multicolumn{2}{c|}{multiclass} & \multicolumn{2}{c|}{Binary Class}       \\ \cmidrule(l){2-5} 
                       & Mean $F1$        & Mean $IoU$        & Mean $F1$ & \multicolumn{1}{c|}{Mean $IoU$} \\ \midrule
LDNet             & \textbf{75.80} & \textbf{62.79} & \textbf{85.18}  & \textbf{76.71}  \\ \midrule
LDNet-VGG16       & 74.42 & 61.16 & 83.75  & 74.98  \\ \midrule
LDNet-ResNet-18    & 73.92 & 60.56 & 84.127 & 75.62  \\ \midrule
LDNet-ResNet-50    & 74.48 & 60.20 & 84.71  & 76.124 \\ \midrule
LDNet-MobileNetv2 & 74.15 & 60.79 & 84.03  & 75.30  \\ \midrule
LDNet-DenseNet    & 74.90 & 61.69 & 84.11  & 75.62  \\ \midrule
LDNet-ShuffleNet  & 72.72 & 59.17 & 83.52  & 74.70  \\ \bottomrule
\end{tabular}%
}
\end{table}
Fig. \ref{atten-map} shows the visualization of the feature activations. The output of the LDNet is a feature map with 5 layers. Each layer predicts 5 classes, four labels corresponding to different lanes and the background. The orange color shows the class predicted in each image. The first row from left shows the predicted background, lane-1, lane-2, lane-3 and lane-4. The blue color shows the remaining pixels.  
The comparison between using an attention-guided decoder with a convolution decoder is illustrated. The attention-guided decoder shows improved localization of features, which eliminates the need for external localization of the features and postprocessing steps.

\begin{figure}[t]
      \centering
      \includegraphics[width=9cm]{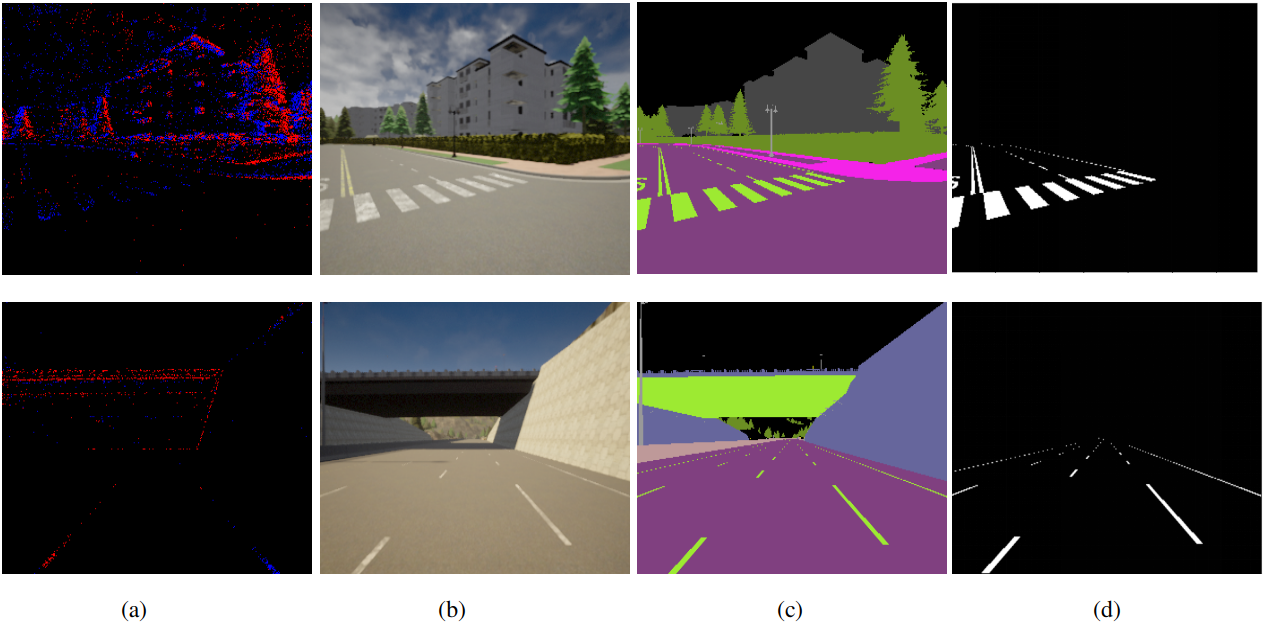}
\caption{The Carla-DVS dataset: (a) shows the event information. The blue and red colors in the event information show the increase and decrease of brightness of that specific pixel, respectively. (b) shows the frame-based RGB image, (c) shows the semantic ground-truth labels, and (d) shows the lane binary labels obtained from semantic labels.  } 
      \label{carla}
\end{figure}

\par

\begin{table}[]
\centering
\caption{Quantitative analysis illustrating the effect of dropout and the DropBlock on LDNet. The evaluation is performed for both multiclass and binary class labels, and the mean $F1$ score and $IoU$ are evaluated for each case and label.}
\label{table 4}
\resizebox{8cm}{!}{%
\begin{tabular}{@{}c|c|c|c|c@{}}
\toprule
\multirow{2}{*}{Model} & \multicolumn{2}{c|}{multiclass} & \multicolumn{2}{c|}{Binary Class} \\ \cmidrule(l){2-5} 
                   & Mean $F1$ & Mean $IoU$ & Mean $F1$ & \multicolumn{1}{c|}{Mean $IoU$} \\ \midrule
LDNet-no dropout & 74.36   & 61.09    &    84.17     &   75.70                            \\ \midrule
LDNet-dropout2d  & 72.47   & 58.94    &    83.25     &      72.80                         \\ \midrule
LDNet-DropBlock  & \textbf{75.80}   & \textbf{62.79}    & \textbf{85.18}   & \textbf{76.71}                         \\ \bottomrule
\end{tabular}%
}
\end{table}

\begin{figure*}[t]
      \centering
      \includegraphics[width=18cm]{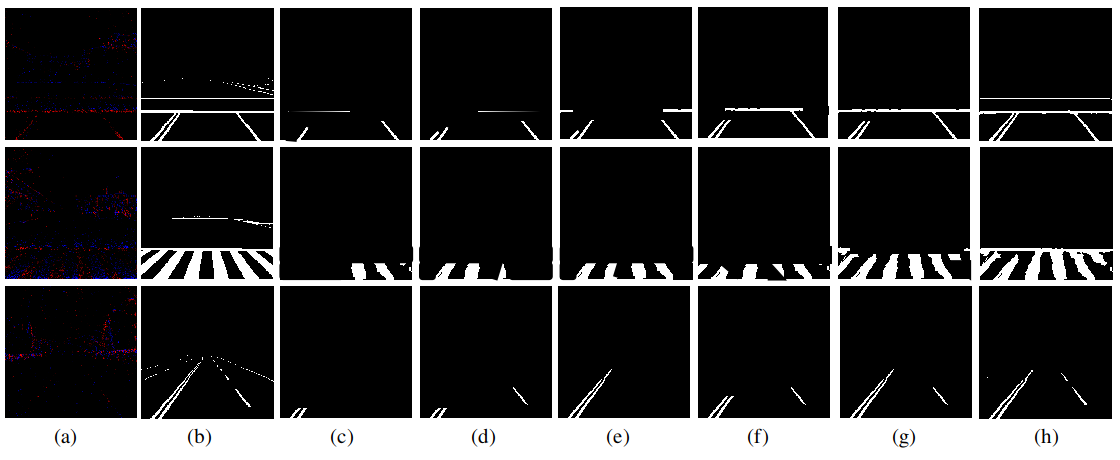}
\caption{Qualitative comparison between different semantic segmentation methods on Carla dataset (a) shows the input image. (b) shows the ground-truth image. (c-h) show the results for FCN, DeepLabv3, RefineNet, SCNN, LaneNet and LDNet (ours), respectively.}
      \label{carla-result}
\end{figure*}

\begin{figure*}[t]
      \centering
      \includegraphics[width=18cm]{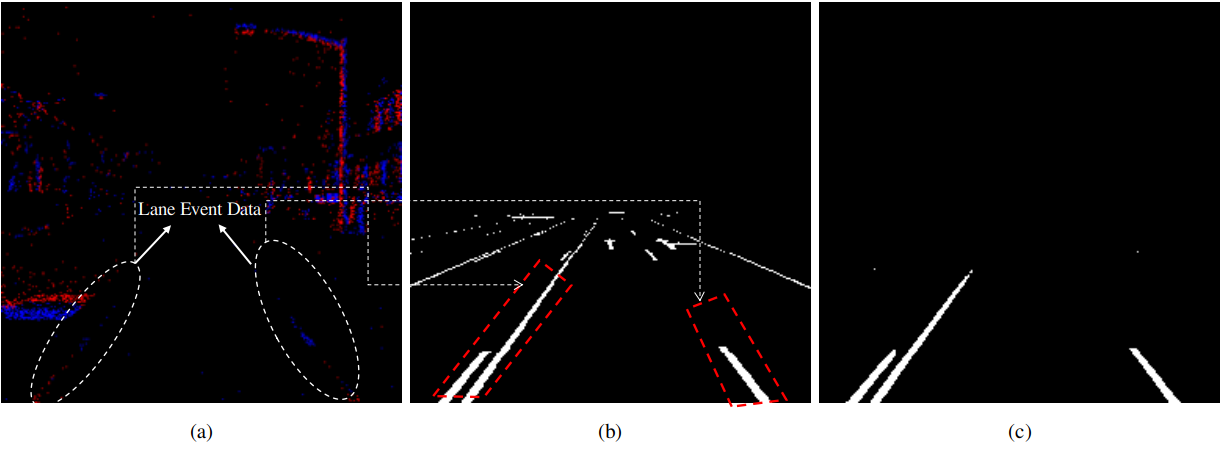}
\caption{Illustrations of the case-study of having no predictions by the proposed network in some regions. (a) explains the lane event dataset is available near the simulated vehicle. (b) shows the projection visualization that is learned by the network. (c) illustrates the prediction results by the proposed LDNet.}
      \label{fail}
\end{figure*}

\section{Ablation Study}

\begin{table}[]
\centering
\caption{Comparison of the evaluation results of LDNet with other state-of-the-art methods on the Carla-DVS dataset. The mean $F1$ scores ($\%$) and mean $IoU$s ($\%$) are used as evaluation metrics for the binary class labels. The values in bold are the best scores.}
\label{table-5}
\resizebox{8.5cm}{!}{%
\begin{tabular}{@{}ccc@{}}
\toprule
Model                                    & Mean $F1$ (\%)                    & Mean $IoU$ (\%) \\ \midrule
\multicolumn{1}{c|}{FCN}       & \multicolumn{1}{c|}{46.15} & 41.32    \\ \midrule
\multicolumn{1}{c|}{DeepLabv3 } & \multicolumn{1}{c|}{50.42} & 42.10    \\ \midrule
\multicolumn{1}{c|}{RefineNet} & \multicolumn{1}{c|}{54.93} & 45.34    \\ \midrule
\multicolumn{1}{c|}{LaneNet }   & \multicolumn{1}{c|}{58.25} & 49.81   \\ \midrule
\multicolumn{1}{c|}{SCNN }      & \multicolumn{1}{c|}{59.15} & 53.14    \\ \midrule
\multicolumn{1}{c|}{\textbf{LDNet-binary class (ours)}}  & \multicolumn{1}{c|}{\textbf{63.50}} & \textbf{58.45} \\ \midrule
\end{tabular}%
}
\end{table}

\begin{figure*}[t]
      \centering
      \includegraphics[width=18cm]{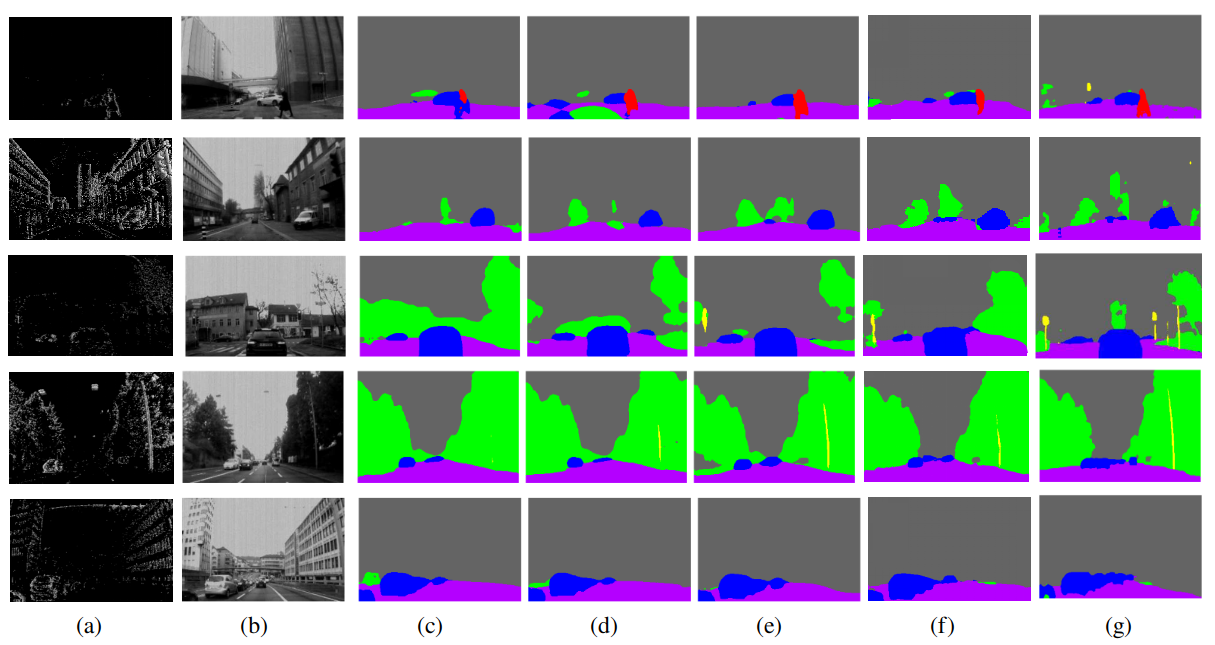}
\caption{The qualitative comparison between different semantic segmentation methods on the EV-Seg dataset, where (a) shows the input image and (b) shows the grayscale image. (c-g) show the results for Basic Dense encoding, Temporal dense encoding, EV-SegNet LDNet (ours) and the ground truth, respectively. In the semantic map, the blue color corresponds to the vehicle, the red color to a person, the green color to vegetation, the purple color to roads, the yellow color to poles and lamps and gray to the sky and buildings. }
      \label{EV-result}
\end{figure*}

\subsection{Perfomance of LDNet on Carla-DVS Data}
For the proposed LDNet method's efficacy, a synthetic dataset using the Carla open-source driving simulator is utilized \cite{evd}. The dataset consists of event data, semantic labels and frame-based RGB images. Fig.\ref{carla} illustrates a sample of data that is used for the evaluation of LDNet. In the dataset, all the map data having different weather conditions are utilized. Furthermore, we sampled the data that contain the lane information for the evaluation of LDNet. The training details for LDNet are similar to the descriptions in section \ref{marker} with training samples of $2522$ and test samples of $1220$, respectively.
\par 
The quantitative evaluation of the proposed LDNet is performed with the same aforementioned lane detection algorithms on the Carla-DVS dataset. Table \ref{table-5} illustrates the mean F1 and IoU scores of the LDNet along with other state-of-the-art algorithms. Fig.\ref{carla-result} shows the qualitative results of LDNet in comparison to the other algorithms.
\par
The qualitative and quantitative results depict the efficacy of the proposed LDNet method, indicating that it surpasses the state-of-the-art lane detection algorithms. However, as the proposed method is applied on simulated data, it is assumed that the results will be better than a real-world dataset. In contrast, the results could be improved compared to the real-world dataset. To analyze this case-study, we review the dataset and find that the number of event data points is not sufficient at far distances from the simulated vehicle compared to locations near the vehicle. The network can learn the schematic at the front using the available event data points but is limited to no predictions at far distances using the same event data. Fig.\ref{fail} illustrates this behavior for the prediction lanes using the Carla-DVS dataset.

\subsection{Performance of LDNet on Event-Segmentation Data}
We perform extensive experimentation of the proposed method on the dynamic vision sensor. Considering that dynamic vision sensor is a newly evolving sensor, not many public datasets are available. Due to the scarcity of lane detection datasets, we used 
the Event Segmentation dataset
 \cite{EVnet} to test the proposed algorithm's generalization. The Event Segmentation dataset is an extension of the DDD17 dataset\cite{24}, which added semantic segmentation ground truth to the original DDD17 dataset consisting only of grayscale images and event information. The semantic labels have six classes: buildings, objects, trees, person, vehicles and background. The LDNet is trained in an end-to-end manner on the Event Segmentation dataset. The training parameters are similar, as described in section \ref{marker}. For a fair comparison, the dataset has the standard split consisting of 15,950 training event images and 3890 test event images.
\par
Table \ref{EV} shows the quantitative evaluation on the Event Segmentation dataset. We compared our model with already existing algorithms. The $IoU$ and $Accuracy$ show the efficacy of LDNet. All the training and testing data were recorded at $50-ms$ intervals. The proposed algorithm does not perform any encoding or preprocessing of the input data. The attention-guided decoder mechanism helps the network to learn the localization of the features. Fig. \ref{EV-result} shows the qualitative comparison of semantic segmentation.   

\subsection{ Effect of frame-based images on LDNet}
\begin{figure}[t]
\centering
\includegraphics[width=8.5cm]{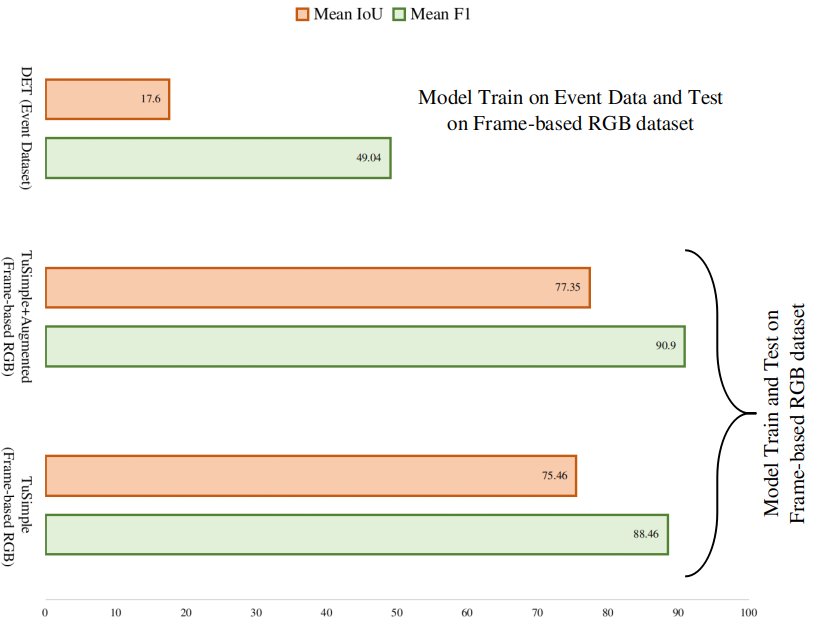}
\caption{ The bar graph shows the quantitative analysis between the frame-based RGB TuSimple dataset and the Event Camera dataset on the LDNet. This graph illustrates the efficacy of the proposed method when trained on the TuSimple dataset. For a fair comparison with the event camera dataset, the proposed network parameters are kept the same in this experimental analysis for the frame-based RGB TuSimple dataset. }
\label{tusimple-r3}
\end{figure}

We experimented with the effect of frame-based images on the proposed LDNet. To validate this claim, we performed experiments with LDNet on the TuSimple dataset. The TuSimple dataset provides the RGB images with the corresponding lane labels. The dataset has 3626 training images and 2782 testing images. 
First, we evaluated LDNet trained on event camera images with TuSimple testing images. Afterward,  
we trained the LDNet on the TuSimple dataset to make a fair analysis for the lane marking detection task. We did not optimize the network parameters for the TuSimple dataset, and the network is used in the same configuration as optimized on the event camera dataset.  Fig. \ref{tusimple-r3} illustrates the quantitative results of LDNet with TuSimple. The environmental conditions covered in the TuSimple dataset are limited; therefore, we also trained the LDNet with the augmented TuSimple dataset. The augmentations on the TuSimple dataset are sun glare, illumination variations and motion blur. Moreover, $30\%$ of the images in the training dataset were augmented. The evaluation of the TuSimple test data result is shown in Fig. \ref{tusimple-r3}.

\begin{table}[t]
\centering
\caption{Comparison of the evaluation results of LDNet with other state-of-the-art methods on the Event Segmentation dataset. The mean F1 scores (\%), mean IoUs (\%) and accuracy are used as evaluation metrics for the semantic labels. ``-`` indicates the metric is not included in the evaluation. The values in bold are the best scores.}
\label{EV}
\resizebox{9cm}{!}{%
\begin{tabular}{@{}l|l|c|c@{}}
\toprule
Model                                              & Accuracy       & Mean $F1$ (\%) & Mean $IoU$ (\%)       \\ \midrule
Basic Dense encoding \cite{24}    & 88.85          & -       & 53.07          \\
Temporal dense encoding \cite{36} & 88.99          & -       & 52.32          \\
EV-SegNet \cite{EVnet}            & 89.76          & -       & 54.81          \\
LDNet (our)                                        & \textbf{90.12} & 69.09   & \textbf{58.12} \\ \bottomrule
\end{tabular}%
}
\end{table}

\section{Conclusion}
Both academia and industry have spent considerable resources and efforts to bring autonomous driving closer to real-world applications. The main challenge is to design reliable algorithms that work in diverse environmental scenarios. There has been extensive development at the algorithm level inspired by deep neural networks. Furthermore, the new sensor development work is progressing, and deployment in the autonomous driving sensor suite continues to mature. The sensor setup might be redundant, but different sensor modalities complement each other to achieve the safety of the autonomous vehicle. Event cameras are fast-growing sensors that provide information with precise timing. In contrast to the event cameras, frame-based cameras and Lidar are sampling-based sensors that oversample distant structures and undersample close structures. Moreover, event cameras capture the scene with precise timing when there is a change in brightness. Thus, they provide a very high dynamic range and low latency compared to standard conventional sensors. 
\par
In this paper, we proposed LDNet, a novel encoder-decoder architecture for lane marking detection in event camera images. LDNet extracts higher-dimensional features from an image, refining full-resolution detections. We introduced the ASPP block as the core of the network, which increases the respective field of the feature map without increasing the number of training parameters. The use of an attention-guided decoder improves the localization of features in the feature map, hence removing the need for the postprocessing step. The proposed network was evaluated on an event camera benchmark, and it was found to outperform the best-performing state-of-the-art methods in terms of the mean $F1$ and $IoU$ scores. LDNet achieves mean $F1$ scores of $75.58\%$ and $85.13\%$ and mean $IoU$s of $62.79\%$ and $76.71 \%$ for multiclass and binary-class tasks, respectively. Moreover, an ablation study is performed on two datasets, i.e. the Carla-DVS dataset and Event Segmentation dataset, which shows the efficacy of LDNet.
\par
The utilization of an event camera in contrast to a frame-based camera is beneficial for the autonomous vehicle's perception of the environment because the event camera dataset is invariant to illumination conditions. In future work, one possible direction is to investigate the application of the current work with the planning and control module of autonomous driving \cite{swn}\cite{sa} for lane keeping and lane changing tasks.

\ifCLASSOPTIONcaptionsoff
\newpage
\fi




\bibliographystyle{IEEEtran}
\bibliography{ref}

\newpage 
\begin{IEEEbiography}[{\includegraphics[width=1in,height=1.25in,clip,keepaspectratio]{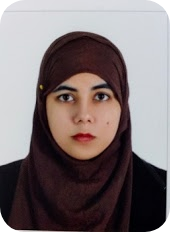}}]{Farzeen Munir} received a B.S. degree in Electrical Engineering from Pakistan Institute of Engineering and Applied Sciences, Pakistan in 2013,
and an M.S. degree in System Engineering from Pakistan Institute of Engineering
and Applied Sciences, Pakistan in 2015. She is currently pursuing a PhD
at Gwangju Institute of Science and Technology, Korea in Electrical
Engineering and Computer Science. Her current research interest include,
machine Learning, deep neural network, autonomous driving and computer vision.
\end{IEEEbiography}

\begin{IEEEbiography}[{\includegraphics[width=1in,height=1.25in,clip,keepaspectratio]{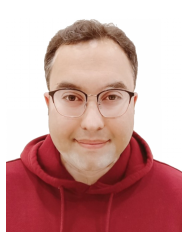}}]{Shoaib Azam} received a B.S. degree in Engineering Sciences from Ghulam Ishaq Khan Institute of Science and Technology, Pakistan, in 2010, and an M.S. degree in Robotics and Intelligent Machine Engineering from National
University of Science and Technology, Pakistan, in 2015. He is currently
pursuing a Ph.D. with the Department of Electrical Engineering and
Computer Science, Gwangju Institute of Science and Technology, Gwangju,
South Korea. His current research interests include deep learning and autonomous driving.
\end{IEEEbiography}


\begin{IEEEbiography}[{\includegraphics[width=1in,height=1.25in,clip,keepaspectratio]{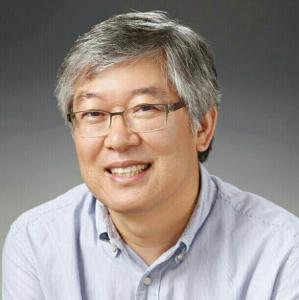}}]{Moongu Jeon} received a B.S. degree in Architectural Engineering from Korea University, Seoul, South Korea, in 1988, and M.S. and Ph.D.
degrees in Computer Science and Scientific Computation from the University
of Minnesota, Minneapolis, MN, USA, in 1999 and 2001, respectively. As
a Post-Graduate Researcher, he worked on optimal control problems at the
University of California at Santa Barbara, Santa Barbara, CA, USA, from
2001 to 2003, and then moved to the National Research Council of Canada,
where he worked on the sparse representation of high-dimensional data and
image processing until 2005. In 2005, he joined the Gwangju Institute of
Science and Technology, Gwangju, South Korea, where he is currently a Full
Professor with the School of Electrical Engineering and Computer Science.
His current research interests are in machine learning, computer vision, and
artificial intelligence.
\end{IEEEbiography}

\begin{IEEEbiography}[{\includegraphics[width=1in,height=1.25in,clip,keepaspectratio]{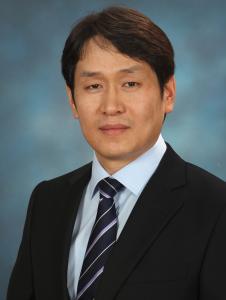}}]{Byung-Geun Lee } received a B.S. degree in Electrical Engineering from Korea University, Seoul, South Korea, in 2000, and M.S. and Ph.D. degrees in Electrical and Computer Engineering from The University of Texas at Austin, Austin, TX, USA, in 2004 and 2007, respectively. From 2008 to 2010, he was a Senior Design Engineer with Qualcomm, San Diego, CA, USA, where he was involved in the development of various mixed-signal ICs. Since 2010, he has been with the Gwangju Institute of Science and Technology (GIST), Gwangju, South Korea. He is currently an Associate Professor with the School of Electrical Engineering and Computer Science, GIST. His research interests include high-speed data converters, CMOS image sensors, and neuromorphic system design.
\end{IEEEbiography}

\begin{IEEEbiography}[{\includegraphics[width=1in,height=1.25in,clip,keepaspectratio]{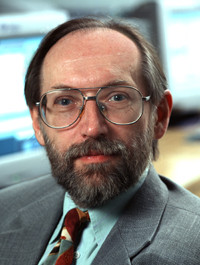}}]{Witold Pedrycz}
received M.Sc., Ph.D., and
D.Sc. degrees from the Silesian University of Technology, Gliwice, Poland.
He is a Professor and the Canada Research Chair
of Computational Intelligence with the Department
of Electrical and Computer Engineering, University
of Alberta, Edmonton, AB, Canada. He is also affiliated with
the Systems Research Institute, Polish Academy of
Sciences, Warsaw, Poland. Dr. Pedrycz is a Foreign
Member of the Polish Academy of Sciences and
a fellow of the Royal Society of Canada. He has
authored 17 research monographs and edited volumes covering various aspects
of computational intelligence, data mining, and software engineering. His
current research interests include computational intelligence, fuzzy modeling
and granular computing, knowledge discovery and data science, fuzzy control,
pattern recognition, knowledge-based neural networks, relational computing,
and software engineering.
Dr. Pedrycz was a recipient of the Prestigious Norbert Wiener Award from
the IEEE Systems, Man, and Cybernetics Society in 2007; the IEEE Canada
Computer Engineering Medal; the Cajastur Prize for Soft Computing from the
European Centre for Soft Computing; the Killam Prize; and the Fuzzy Pioneer
Award from the IEEE Computational Intelligence Society. He is vigorously
involved in editorial activities. He is an Editor-in-Chief of Information
Sciences, Editor-in-Chief of WIREs Data Mining and Knowledge Discovery
(Wiley), and the International Journal of Granular Computing (Springer). He
currently serves on the Advisory Board of IEEE Transactions on Fuzzy
Systems and is a member of a number of editorial boards of other international
journals
\end{IEEEbiography}



\end{document}